\documentclass[11pt,a4paper]{article}

\usepackage[T1]{fontenc}
\usepackage[utf8]{inputenc}
\usepackage[a4paper,margin=1in]{geometry}
\usepackage{lmodern}
\usepackage{microtype}
\usepackage{amsmath,amssymb,amsfonts}
\usepackage{algorithmic}
\usepackage{graphicx}
\usepackage{textcomp}
\usepackage{multirow}
\usepackage{booktabs}
\usepackage{caption}
\usepackage[numbers,sort&compress]{natbib}
\usepackage[colorlinks=true,linkcolor=blue,citecolor=blue,urlcolor=blue]{hyperref}

\hypersetup{
  pdftitle={ARGUS: Accelerated, Robust, General, and Unsupervised Cell Tracking Solutions},
  pdfauthor={Noah Jaitner, Kandice Tanner, Ingolf Sack, Hossein S. Aghamiry}
}

\begin{document}

\title{\vspace{-1.2em}\bfseries ARGUS: Accelerated, Robust, General, and Unsupervised Cell Tracking Solutions}

\author{
Noah Jaitner\textsuperscript{1}
\quad Kandice Tanner\textsuperscript{2}
\quad Ingolf Sack\textsuperscript{1}
\quad Hossein S.\ Aghamiry\textsuperscript{1,*}\\[0.75em]
\small \textsuperscript{1}Department of Radiology, Charité -- Universitätsmedizin Berlin, 10117 Berlin, Germany\\
\small \textsuperscript{2}Laboratory of Cell Biology, Center for Cancer Research,\\ \small{National Cancer Institute, National Institutes of Health, Bethesda, MD, USA}\\
\small \textsuperscript{*}Corresponding author: \href{mailto:hossein.aghamiry@charite.de}{hossein.aghamiry@charite.de}
}

\date{}
\maketitle

\begin{abstract}
\textbf{Background and Objective:} Quantitative analysis of cell dynamics is central to modern biological research, providing critical insights into immune cell interactions, disease progression, and drug mechanisms. Automated cell tracking in time-lapse microscopy remains challenging due to noise, morphological variations, overlapping cells, and dynamic events such as divisions and fusions.

\textbf{Methods:} We present ARGUS, a framework for \textit{A}ccelerated, \textit{R}obust, \textit{G}eneral, and \textit{U}nsupervised cell tracking \textit{S}olutions. ARGUS combines adaptive cell detection, dense Farneback optical-flow prediction, frame-to-frame linear assignment, and a sequence-level tracklet-refinement step that reconnects trajectory fragments across short temporal gaps.

\textbf{Results:} On publicly available Cell Tracking Challenge datasets, ARGUS achieved detection accuracy of 0.905--0.971 and tracking accuracy of 0.897--0.964, with runtimes within 1 minute (5--6 seconds for 3 frames).

\textbf{Conclusions:} ARGUS is a modular, interpretable framework that can be adapted to different imaging modalities and biological applications without training data or GPU infrastructure. The implementation is publicly available at \url{https://github.com/Gitinc/argus}.
\end{abstract}

\noindent\textbf{Keywords:} Cell tracking; Optical flow; Monogenic signal; Assignment problem; Tracklet association; Time-lapse microscopy
\vspace{1em}

\section{Introduction}

Live-cell imaging combined with quantitative cell tracking has become essential for studying cellular processes. By tracking cells over time, researchers can directly observe migration, proliferation, differentiation, and apoptosis \cite{elifesciences}. They can also characterize how cells interact and respond to therapeutic compounds \cite{elifesciences, wang2024cart, mcswiggen2024, victora2024}. From a clinical perspective, these measurements are important for CAR-T cell therapy \cite{wang2024cart}, high-throughput drug screening \cite{mcswiggen2024}, and detection of circulating tumor cells \cite{cellreports2024, scirep2024}. Development of automated, large-scale spatiotemporal imaging methods allows millions of cells to be imaged per day at subcellular resolution \cite{ctc2023}. However, the resulting terabyte-scale datasets require efficient computational methods that can extract reliable quantitative information \cite{elifesciences, frontiersin}.

2D live-cell microscopy tracking remains computationally challenging. The Cell Tracking Challenge (CTC), an ongoing community benchmark, has assessed algorithmic progress over the past decade \cite{ctc2023, ctc2017, ctc2014}. No method has yet achieved perfect accuracy across all datasets. Reported scores range from 52\% to 99\%, depending on dataset complexity \cite{ctc2023}. These challenges arise from image noise, defocused cells, morphological variability, overlapping cells, cell division, fusion, and data volumes that often exceed 500 GB.

Existing tracking methods can be broadly grouped into local and global approaches. Local methods, such as nearest-neighbor linking, are computationally fast but prone to errors during occlusion and division. In such cases, accuracy can decrease by 20-30\% \cite{ctc2023}. Global methods consider the full image sequence and usually achieve higher accuracy through three main optimization strategies:
\begin{itemize}
    \item \textbf{Dynamic Programming}: The Viterbi algorithm and related methods \cite{viterbi} maximize a scoring function over event probabilities across the full sequence. These approaches can produce globally optimal tracks, but their complexity often scales quadratically with the number of detections. Notably, the KTH-SE algorithm \cite{magnusson2014global} showed that a classical Viterbi formulation, without machine learning, remains competitive at the top of the CTC leaderboard \cite{ctc2023}.
    
    \item \textbf{Integer Linear Programming (ILP)}: Binary-variable formulations \cite{ilp} jointly optimize the full spatiotemporal context and can handle complex events such as division and fusion. However, they face two practical limitations. First, their worst-case complexity is exponential and often requires GPU acceleration for large datasets \cite{ctc2023}. Second, they rely on hand-crafted cost functions. Recent transformer-based methods such as Trackastra \cite{trackastra2024} address the second limitation by learning cost functions from data.
   
    \item \textbf{Minimum-Cost Flow}: Tracking is formulated as a flow-network problem \cite{mincostflow}. This approach handles cell division robustly, but it requires careful cost-function design and may not scale to very large datasets. The KIT-GE(3) method \cite{scherr2020cell} combines minimum-cost flow with a dual-branch U-Net for dense cell populations \cite{ctc2023}.
\end{itemize}

Optical flow-based methods represent an intermediate position. They estimate dense motion fields between consecutive frames, which reduces the association search space from $O(n^2)$ to $O(n)$ per frame pair \cite{opticalflow1, opticalflow2, farneback}. These methods can match the accuracy of global methods while running an order of magnitude faster. Complex events such as mitosis and extended occlusions remain challenging. However, their speed makes them attractive for practical use.

We present ARGUS, a framework for \textit{A}ccelerated, \textit{R}obust, \textit{G}eneral, and \textit{U}nsupervised cell tracking \textit{S}olutions. ARGUS uses a two-stage association strategy. Dense Farneback optical-flow prediction and frame-to-frame linear assignment generate initial trajectory fragments. A subsequent sequence-level refinement step then reconnects compatible fragments across short temporal gaps. This design follows established two-stage tracking principles \cite{jaqaman2008} and avoids the exponential complexity of joint sequence-level optimization.

ARGUS comprises four components:
\begin{enumerate}
    \item \textbf{Data preparation and cell detection:} Locate candidate cells in each image frame.
    \item \textbf{Motion estimation:} Estimate inter-frame displacement using dense optical flow.
    \item \textbf{Local association:} Link detections between consecutive frames by linear assignment in motion-corrected space.
    \item \textbf{Global tracklet refinement:} Reconnect compatible trajectory fragments across short temporal gaps.
\end{enumerate}

The methodological core of the linking step is the use of a dense optical-flow field as a spatially varying motion prior. Unlike proximity-based or constant-velocity association schemes, the flow field provides a per-pixel displacement estimate. This estimate adapts to locally heterogeneous cell motion. As a result, different regions of the field of view can receive different predicted displacements within the same frame. The framework is training-free and does not require GPU infrastructure. This makes it suitable for exploratory studies and real-time monitoring when annotated data are unavailable. The paper is organized as follows. Section~2 presents the optimization formulation and the four processing stages. Section~3 reports quantitative and qualitative results on four CTC benchmark datasets. Section~4 discusses strengths, limitations, and potential extensions.

\section{Methodology}

We formulate ARGUS as a tracking-by-detection framework in which inter-frame association and sequence-level refinement are solved separately. Let $D_t = \{d_{t,1}, \dots, d_{t,n_t}\}$ be the detected cell centroids in frame $t$, with $d_{t,i}\in\mathbb{R}^2$. For each detection $d_{t,i}$, an optical-flow displacement $v_{t,i}$ predicts its next position $p_{t+1,i}=d_{t,i}+v_{t,i}$. Binary variables $x^t_{ij}\in\{0,1\}$ indicate whether $d_{t,i}$ is linked to $d_{t+1,j}$. For each consecutive frame pair, ARGUS solves
\begin{equation}
\min_{x^t_{ij}} \sum_{i=1}^{n_t} \sum_{j=1}^{n_{t+1}} c^t_{ij}x^t_{ij},
 \text{~~where~~}
c^t_{ij}=\|p_{t+1,i}-d_{t+1,j}\|_2,
\end{equation}
subject to one-to-one constraints. Here, $\|\cdot\|_2$ denotes the Euclidean norm. Candidate links with $c^t_{ij}>\delta$ are forbidden. This local optimization produces initial trajectory fragments. A second sequence-level optimization reconnects compatible fragments, as described below. The resulting decomposition avoids a computationally expensive joint optimization over all detections while improving temporal consistency.

\begin{figure}[ht!]
\centerline{\includegraphics[width=\textwidth]{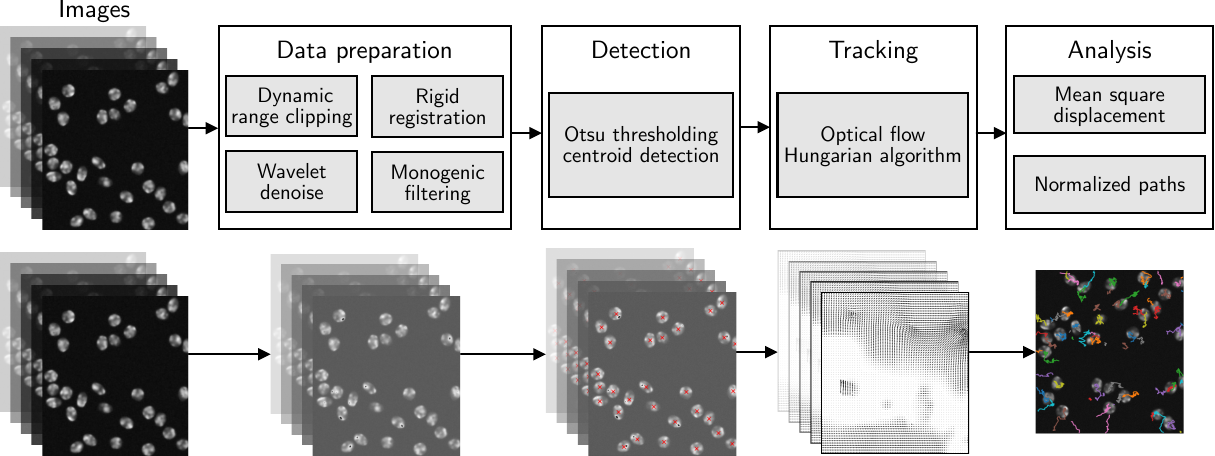}}
\caption{Overview of the proposed processing framework. Images are preprocessed and cells are detected using thresholding and centroid extraction. Initial trajectories are generated by optical-flow-guided frame-to-frame assignment and subsequently refined by global tracklet association across short gaps. The resulting trajectories are used for migration analysis.}
\label{fig1}    
\end{figure}

\subsection{Data Preparation}

Microscopy images often exhibit low contrast, uneven illumination, and high noise levels, which hinder reliable cell detection. To address these issues, ARGUS includes a data-preparation stage before detection. Cell contrast enhancement by percentile-based intensity clipping and wavelet denoising were applied to every dataset as standard operations because the ablation study showed that they are necessary for reliable detection (Table~\ref{tab:res}). Additional operations are imaging-modality specific rather than dataset specific: monogenic phase-symmetry filtering was used for phase-contrast data, where halo artifacts and weak cell contrast hinder segmentation, and drift correction can be inserted for time-lapse data affected by stage motion. The technical details of these data-preparation modules, including the exact contrast-enhancement and wavelet-denoising settings, are provided in \ref{app:data_prep}.

\subsection{Cell Detection}

Preprocessed cell images are converted into a binary segmentation mask by adaptive thresholding. In practice, the threshold is chosen via Otsu's method \cite{otsu}. Let $S_t(x)$ denote the preprocessed intensity at pixel $x$ in frame $t$. All pixel values with $S_t(x)$ above the threshold are marked as foreground (candidate cell regions), and others as background. Very small connected components (below the minimum cell size) are further removed as likely noise artifacts. The minimum cell size is set by the operator as described below. The output is a set of segmented blobs, each representing a cell or cluster of cells in that frame.

The centroid of each detected cell is then computed. We enforce a minimum separation between detected centroids, on the order of the typical cell radius, to stabilize later linking procedures and minimize cell detections that are too close. This cell-size parameter (Table~\ref{tab:params_runtime}) is the single operator-specified length scale in ARGUS: it sets the minimum connected-component size described above, the minimum centroid separation, and, as described below, the gating radius $\delta$ used for motion-based association and the distance threshold for division detection. Each detected centroid is recorded as a cell detection point $d_{t,i}$ for that frame. This forms the input to the tracking stage.

\subsection{Motion Estimation and Tracking}

With cells detected in each frame, the next stage is to link these detections over time into trajectories. We integrate optical flow-based motion estimation \cite{Guo_Ven_Zhou_2013} with a frame-to-frame linear assignment algorithm. The optical flow provides a \emph{directional prediction} where each cell will move in the next frame, thereby reducing ambiguity in the association step. Specifically, we employ the Farneback optical flow \cite{farneback} method, which estimates a dense flow field by approximating the local image neighborhood with polynomial expansions. For each pixel, the intensity structure is modeled by a quadratic polynomial, and the displacement field is obtained by matching these polynomial representations between consecutive frames. This choice is motivated by the need to handle the smooth, continuous motion typical in fluorescence microscopy while remaining computationally efficient. The Farneback model assumes that the flow field $u_t$ can be recovered from the displacement of polynomial basis functions fitted to $I_t$ and $I_{t+1}$, leading to a weighted least-squares problem of the form
\begin{equation}
    u_t(x)=\arg\min_{\eta\in\mathbb{R}^2} \sum_{\Delta x \in \mathcal{N}} w(\Delta x) \left\| A(x + \Delta x)\, \eta - \Delta b_t(x + \Delta x) \right\|^2_2,
\end{equation}
where $\mathcal{N}$ is a local spatial neighborhood around pixel $x$, $A(x + \Delta x)$ encodes the second-order polynomial coefficients of the local intensity structure, $\Delta b_t(x + \Delta x)$ encodes the temporal intensity difference between frames $t$ and $t+1$, and $w(\Delta x)$ is a weight function that emphasizes nearby pixels. The flow is estimated in a coarse-to-fine pyramid scheme \cite{farneback}, which extends the capture range to displacements larger than a single pixel neighborhood. We used a three-level pyramid with a scale factor of 0.5 between levels, a 15-pixel averaging window, three iterations per level, and a second-order polynomial expansion over a neighborhood of size 5 smoothed with a Gaussian of standard deviation 1.2; these settings were kept identical across all datasets. Once the optical flow $u_t$ is computed, we predict the next position of each cell by sampling the flow at the cell location. In other words, for each detected point $d_{t,i} = (x,y)$, we set $v_{t,i}=u_t(d_{t,i})=(v_{x,t,i},\,v_{y,t,i})$ and $p_{t+1,i}=d_{t,i}+v_{t,i}=(x+v_{x,t,i},\,y+v_{y,t,i})$. This gives a predicted location $p_{t+1,i}$ in frame $t+1$ where cell $i$ from frame $t$ is expected to appear.

Using these predicted positions $P_{t+1} = \{p_{t+1,i}\}$ and the actual detections $D_{t+1} = \{d_{t+1,j}\}$ in the next frame, we formulate data association as a bipartite matching problem. We construct a cost matrix $C^t=[c^t_{ij}]$ for all predicted-actual pairs, where $c^t_{ij}=\|p_{t+1,i} - d_{t+1,j}\|_2$ if the distance is below a threshold $\delta$, and $c^t_{ij}=\infty$ otherwise (effectively forbidding unrealistic matches). The gating radius is set to the cell radius defined above, encoding the prior that a cell cannot displace by more than its own size between consecutive frames. The optimal assignment that minimizes total cost $\sum_{i,j} c^t_{ij}x^t_{ij}$ subject to one-to-one pairing is obtained using a linear assignment solver \cite{College_1955}, yielding the matching set $M_t = \{(i,j)\}$ that indicates which detection in frame $t$ connects to which detection in frame $t+1$. By chaining these assignments over all frames, each cell is tracked along a path through the video.

To handle the birth and death of tracks, we employ simple track management heuristics. Any detection in frame $t+1$ that remains unmatched after the frame-to-frame assignment is initialized as a new track, accounting for cells that newly enter the field of view or become newly segmented. Conversely, if a predicted position from an existing track cannot be matched to any detection, or if multiple tracks are predicted into the same mask label (a collision), the track is marked as occluded rather than immediately terminated. An occluded track is propagated forward using the optical flow field alone for up to five frames, allowing it to re-engage with a detection if the cell reappears within that window.

Cell divisions are handled explicitly. After the frame-to-frame assignment, each matched track is tested for the presence of a nearby unmatched detection lying in a different mask label within a distance of two cell radii. If such a partner is found, a mitosis-like split is triggered: the parent track's tentative extension into frame $t+1$ is rolled back, the parent is marked inactive, and two new daughter tracks are spawned from the matched detection and the partner detection respectively. The parent record retains the identifiers of both daughters, preserving the lineage graph.

\subsection{Global Tracklet Refinement}
Frame-to-frame association may split a trajectory when a cell is temporarily occluded or missed by the detector. Following the two-stage association principle of Jaqaman et al.~\cite{jaqaman2008}, the initial trajectories are treated as tracklets $\{\tau_k\}_{k=1}^{K}$. Let $p_k^{\mathrm{start}}$ and $p_k^{\mathrm{end}}$ denote the first and last centroid positions of tracklet $\tau_k$, respectively. For a candidate link from tracklet $\tau_i$ to a later tracklet $\tau_j$, the temporal gap is $g_{ij}=t_j^{\mathrm{start}}-t_i^{\mathrm{end}}-1$. The endpoint $p_i^{\mathrm{end}}$ is propagated to $t_j^{\mathrm{start}}$ by sequentially applying the intervening optical-flow fields, yielding $\widehat{p}_{ij}$. Candidate links are retained only if $1\leq g_{ij}\leq G_{\max}$ and
\begin{equation}
\|\widehat{p}_{ij}-p_j^{\mathrm{start}}\|_2 \leq \delta\sqrt{g_{ij}+1}.
\end{equation}
For each admissible pair, the refinement cost is
\begin{equation}
C_{ij}=\frac{\|\widehat{p}_{ij}-p_j^{\mathrm{start}}\|_2}{\delta\sqrt{g_{ij}+1}}
+\lambda_g g_{ij}+\lambda_{\theta}(1-\widehat{v}_i^{\mathsf{T}}\widehat{v}_j),
\end{equation}
where $\widehat{v}_i$ and $\widehat{v}_j$ denote the normalized terminal direction of $\tau_i$ and the normalized initial direction of $\tau_j$, respectively, so that the final term penalizes inconsistent motion directions. We used $G_{\max}=5$, $\lambda_g=0.25$, $\lambda_{\theta}=0.25$, and accepted links only when $C_{ij}<2.5$; these values were held constant across datasets. All admissible endpoint links are evaluated jointly in a one-to-one linear assignment, and accepted links are connected into refined trajectories. This stage is global at the tracklet level, rather than a fully global detection-level lineage model; divisions remain handled separately by the split-detection rule described above. Richer branching models, such as minimum-cost-flow or Viterbi formulations, remain appropriate when full lineage reconstruction is required \cite{mincostflow,viterbi}.

\section{Results}

\begin{figure}[h!]
\centerline{\includegraphics[width=\textwidth]{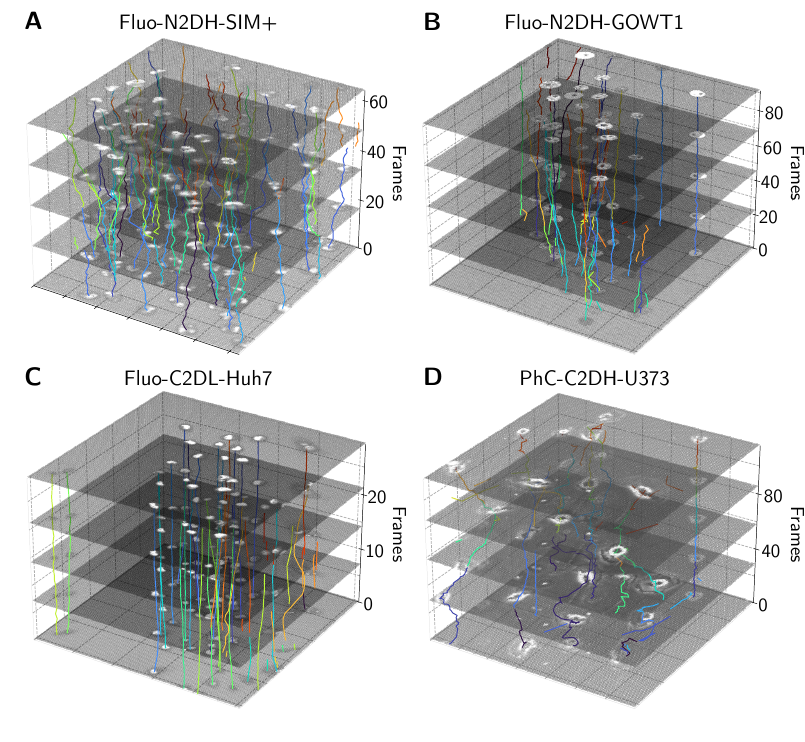}}
\caption{Example of tracked paths for four datasets: (A) Fluo-N2DH-SIM+, (B) Fluo-N2DH-GOWT1, (C) Fluo-C2DL-Huh7 and (D) PhC-C2DH-U373}
\label{fig2}
\end{figure}

The proposed algorithm was tested on four CTC datasets with diverse cell types and imaging modalities: Fluo-N2DH-SIM+ (simulated fluorescence nuclei), Fluo-N2DH-GOWT1 (mouse stem cells), Fluo-C2DL-Huh7 (human hepatocellular carcinoma cells), and PhC-C2DH-U373 (human glioblastoma astrocytes in phase contrast). Performance is quantitatively evaluated using standard CTC metrics summarized in Table \ref{tab:res}, and qualitatively by visual inspection of representative trajectories in Figure \ref{fig2}.

\begin{table}[ht!]
\centering
\caption{Comparative benchmark results for cell tracking. Detection accuracy (DET) and tracking accuracy (TRA) are reported for ARGUS, Trackastra and methods from the Cell Tracking Challenge on the Fluo-C2DL-Huh7, Fluo-N2DH-GOWT1, Fluo-N2DH-SIM+, and PhC-C2DH-U373 datasets \cite{ctc2023}.}
\label{tab:res}
\begin{tabular}{l|cc|cc}
\hline
\multirow{2}{*}{Method}
  & \multicolumn{2}{c|}{Fluo-N2DH-SIM+}
  & \multicolumn{2}{c}{Fluo-N2DH-GOWT1} \\
& DET & TRA & DET & TRA \\ \hline
\textbf{ARGUS}              & 0.971 & 0.964 & 0.943 & 0.937 \\
\quad w/o denoising         & 0.961 & 0.954 & 0.777 & 0.786 \\
\quad w/o clipping          & 0.730 & 0.727 & 0.413 & 0.413 \\
\quad + GT mask             & \textbf{1.000} & \textbf{0.999} & \textbf{1.000} & \textbf{0.999} \\ \hline
\textbf{Trackastra} (GT Mask)   & \textbf{1.000} & \textbf{0.999} & \textbf{1.000} & \textbf{0.999} \\
\quad + ARGUS Mask          & 0.972 & 0.968 & 0.944 & 0.939 \\ \hline
CTC (average)               & 0.967 & 0.961 & 0.887 & 0.873 \\
CTC (best)                  & 0.993 & 0.990 & 0.979 & 0.976 \\
CTC (worst)                 & 0.838 & 0.832 & 0.559 & 0.521 \\ \hline\hline
\multirow{2}{*}{Method}
  & \multicolumn{2}{c|}{Fluo-C2DL-Huh7}
  & \multicolumn{2}{c}{PhC-C2DH-U373} \\
& DET & TRA & DET & TRA \\ \hline
\textbf{ARGUS}              & 0.957 & 0.946 & 0.905 & 0.897 \\
\quad w/o denoising         & 0.957 & 0.946 & 0.905 & 0.897 \\
\quad w/o clipping          & 0.659 & 0.658 & 0.723 & 0.721 \\
\quad + GT mask             & \textbf{1.000} & \textbf{0.999} & \textbf{1.000} & \textbf{0.999} \\ \hline
\textbf{Trackastra} (GT Mask)   & \textbf{1.000} & \textbf{0.999} & \textbf{1.000} & \textbf{0.999} \\
\quad + ARGUS Mask          & 0.957 & 0.946 & 0.905 & 0.898 \\ \hline
CTC (average)               & 0.937 & 0.916 & 0.952 & 0.945 \\
CTC (best)                  & 0.989 & 0.988 & 0.998 & 0.997 \\
CTC (worst)                 & 0.796 & 0.780 & 0.568 & 0.613 \\
\hline
\end{tabular}
\end{table}

The algorithm performed competitively on all datasets, yielding DET scores from 0.905 to 0.971 and TRA scores from 0.897 to 0.964. The highest performance was achieved for Fluo-N2DH-SIM+ (DET: 0.971, TRA: 0.964), followed by Fluo-C2DL-Huh7 (DET: 0.957, TRA: 0.946) and Fluo-N2DH-GOWT1 (DET: 0.943, TRA: 0.937). ARGUS performed above the CTC average on three out of four datasets. The phase-contrast dataset PhC-C2DH-U373 yielded the lowest performance (DET: 0.905, TRA: 0.897), which is nevertheless competitive given the challenges posed by imaging artifacts and lower contrast. An overview of performance is shown in table \ref{tab:res}. 

Data preparation was standardized across datasets. For all datasets, cell contrast was enhanced by percentile-based intensity clipping to compress dynamic range and increase the contrast of low-intensity cells. The clipping percentile and cell-size parameters are shown in Table~\ref{tab:params_runtime}. This was then followed by single-channel wavelet denoising to suppress noise while preserving edges. For PhC-C2DH-U373, where phase-contrast imaging produces characteristic halo artifacts, additional modality-specific monogenic phase-symmetry filtering was performed to enhance circular cell morphology while suppressing edge artifacts.

\begin{table}[t]
  \centering
  \caption{Optimal parameters and computational time per dataset.
           Runtimes are reported for the full dataset and a 3 image subset,
           on the HPC cluster and a standard workstation.}
  \label{tab:params_runtime}
  \begin{tabular}{l c c c c c c}
    \hline
    & & & \multicolumn{4}{c}{Computational time (s)} \\
    \cline{4-7}
    & & & \multicolumn{2}{c}{HPC} & \multicolumn{2}{c}{Workstation} \\
    \cline{4-5} \cline{6-7}
    Dataset & Cell size & Clip thresh.
            & Full & 3 img & Full & 3 img \\
    \hline
    Fluo-N2DH-SIM+ & 30 & 95 & 11 & 5 & 26 & 5\\
    Fluo-N2DH-GOWT1 & 60 & 90 & 17  & 5 & 40 & 6 \\
    Fluo-C2DL-Huh7 & 30 & 97 & 12 & 5 & 25 & 5\\
    PhC-C2DH-U373 & 80 & 97 & 14 & 5 & 46 & 6\\
    \hline
  \end{tabular}
\end{table}

To evaluate the performance of our algorithm we performed an ablation study. The ablation results in Table~\ref{tab:res} reveal the relative contribution of each data-preparation stage. Disabling intensity clipping produced the largest performance drops across all datasets, with DET falling to 0.730 on Fluo-N2DH-SIM+, 0.413 on Fluo-N2DH-GOWT1, 0.659 on Fluo-C2DL-Huh7, and 0.723 on PhC-C2DH-U373, confirming that cell contrast enhancement is the single most critical data-preparation step. Disabling wavelet denoising had a negligible effect on Fluo-C2DL-Huh7 and PhC-C2DH-U373, where noise is less pronounced, but caused a substantial drop on Fluo-N2DH-GOWT1 (DET: 0.777, TRA: 0.786), indicating that wavelet denoising is necessary for datasets with higher background noise and does not degrade performance on lower-noise datasets. The upper-bound experiment using ground-truth segmentation masks yields near-perfect scores (DET: 1.000, TRA: 0.999) on all datasets for ARGUS, demonstrating that the tracking and association components of the pipeline introduce negligible error when provided with accurate detections. 

\begin{figure}[ht]
\centerline{\includegraphics[width=\textwidth]{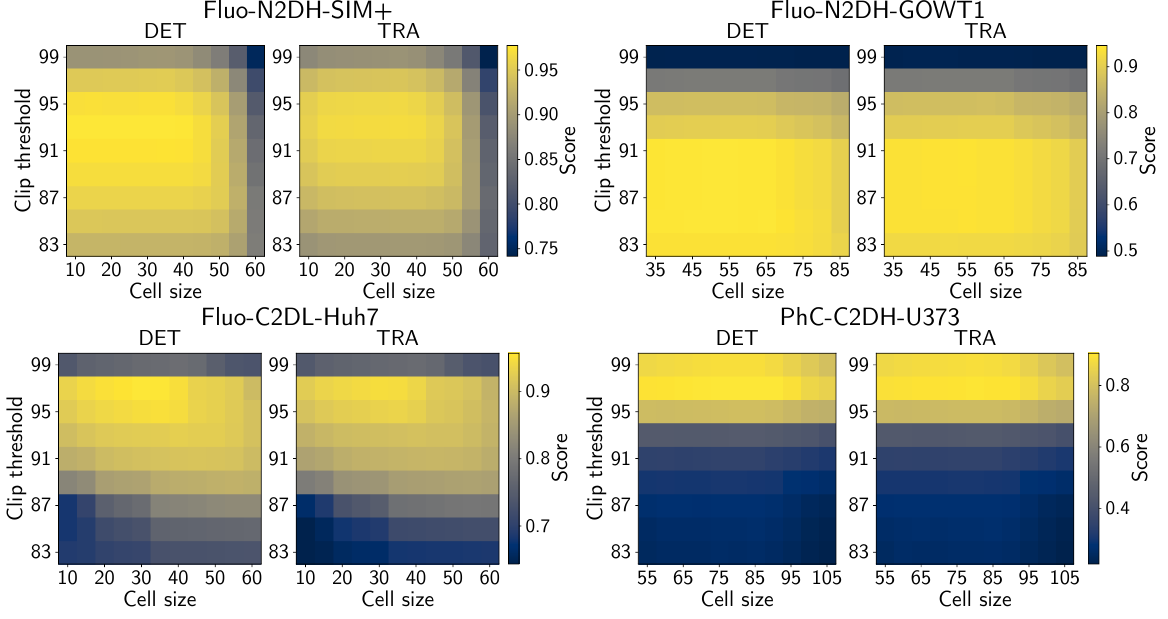}}
\caption{Parameter sensitivity of cell detection and tracking performance across four CTC benchmark datasets. Heatmaps show DET (detection) and TRA (tracking) scores as a function of cell size parameter and clip threshold percentile for Fluo-N2DH-SIM+, Fluo-N2DH-GOWT1, Fluo-C2DL-Huh7, and PhC-C2DH-U373.}
\label{fig:paramscan}
\end{figure}

To assess the sensitivity of ARGUS to the choice of preprocessing parameters, we performed a systematic grid search over the intensity clip threshold percentile (ranging from the 83rd to the 99th percentile) and the cell size parameter across all four benchmark datasets. The results are summarized in Figure~\ref{fig:paramscan}. Performance remained largely stable across a wide range of parameter combinations. For clip thresholds below the 97th percentile, performance degraded by less than 0.05 in DET and TRA on Fluo-N2DH-SIM+ and Fluo-N2DH-GOWT1, and by less than 0.1 on Fluo-C2DL-Huh7 for clip thresholds between 91 and 97. For the PhC-C2DH-U373 dataset performance remained stable for a clip threshold greater than 95. Sensitivity to the cell size parameter was similarly low within a range of approximately $\pm$60\% around the optimum, with performance degrading only at extreme values far outside the typical cell size for each dataset.

Global gap-closing refinement reduced the number of tracks while increasing their mean duration across all four datasets (Table \ref{tab:refinement}), consistent with the linking of fragmented trajectories arising from missed detections or transient occlusions. The track count decreased by 7--14\%, from 127 to 110 for Fluo-N2DH-SIM+, 69 to 64 for Fluo-N2DH-GOWT1, 66 to 61 for Fluo-C2DL-Huh7, and 50 to 43 for PhC-C2DH-U373. The mean track duration increased in every case. These results indicate that the refinement step recovers temporal continuity in trajectories that would otherwise be split into shorter, disconnected segments, yielding fewer but more complete tracks.

\begin{table}[ht]
\centering
\caption{Number of tracks and mean duration in frames before and after global gap-closing refinement.}
\label{tab:refinement}
\begin{tabular}{l|cc|cc}
& \multicolumn{2}{c|}{\textbf{Tracks}} 
& \multicolumn{2}{c}{\textbf{Mean duration}} \\
\textbf{Dataset} 
& \textbf{Pre} & \textbf{Post} 
& \textbf{Pre} & \textbf{Post} \\ \hline
Fluo-N2DH-SIM+ & 127 & 110 & 20.7 & 24.2 \\
Fluo-N2DH-GOWT1 & 69 & 64 & 31.7 & 34.5 \\
Fluo-C2DL-Huh7 & 66 & 61 & 15.6 & 17.1 \\
PhC-C2DH-U373 & 50 & 43 & 22.2 & 26.8 \\
\end{tabular}
\end{table}

We further compared ARGUS to Trackastra. Trackastra is a cell tracking approach that links already segmented cells in a microscopy timelapse by predicting associations with a transformer model \cite{trackastra2024}. We used the general 2D pretrained model. Importantly, Trackastra does not provide any cell detection. When provided with ARGUS-generated masks, Trackastra achieves scores within 0.005 of ARGUS across all four datasets (Table~\ref{tab:res}), and exactly matching scores for Fluo-C2DL-Huh7. When provided with ground-truth masks, both methods reach the same near-perfect performance (DET 1.000, TRA 0.999).

The framework supports parallel processing across CPU cores. Processing was performed on Dell PowerEdge R7525 compute nodes equipped with 12 CPU cores and 120 GB of RAM and on a standard workstation (4 CPU cores and 32 GB of RAM). An overview of processing time is provided in table \ref{tab:params_runtime}. Processing times were below 1 minute per dataset on a standard workstation and below 30 seconds on a high performance cluster, with representative processing speeds of approximately 5 seconds for 3 consecutive frames, enabling real-time monitoring applications.

For context, we also ran Trackastra with its general 2D pretrained model on the same local machine, using ARGUS-generated detections as input. Processing the full sequence took 14 s for Fluo-N2DH-SIM+, 31 s for Fluo-N2DH-GOWT1, 12 s for Fluo-C2DL-Huh7, and 21 s for PhC-C2DH-U373, with approximately 3 s for 3 consecutive frames. While these times are shorter, but the comparison is not directly equivalent: Trackastra requires a pretrained transformer model and, for training, GPU resources, neither of which ARGUS needs, and it performs association only, taking pre-computed segmentation masks as input rather than detecting cells itself. ARGUS therefore provides a complete detection-and-tracking solution at comparable accuracy (Table~\ref{tab:res}) without training data or dedicated hardware.

Qualitative evaluation of trajectories (Figure \ref{fig2}) showed that the algorithm produced smooth, biologically plausible cell trajectories across all datasets. Trajectories demonstrated temporal continuity with no abrupt jumps or erratic switches, confirming successful motion estimation and association. The Farneback optical flow provided good prediction of cell displacements, while the frame-to-frame linear assignment minimized the gated association cost and the tracklet-refinement stage reconnected compatible fragments across short gaps. Track initiations and terminations were properly handled, starting new tracks when cells entered the field of view or appeared by division, while terminating tracks when cells exited or became undetectable. Videos of full tracking results for each dataset are included in supplementary material.

\begin{figure}
\centerline{\includegraphics[width=\textwidth]{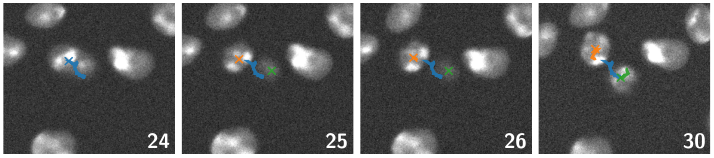}}
\caption{Representative fluorescence microscopy images from Fluo-N2DH-SIM+ dataset showing a single cell undergoing mitosis at frame 25, giving rise to two cells.}
\label{fig:mitosis}
\end{figure}

To qualitatively validate the mitosis detection capability of ARGUS, we examined a representative division event identified in the tracking output. Figure~\ref{fig:mitosis} shows four consecutive frames spanning the division: the mother cell is visible as a single tracked object in frame 24, undergoes division at frame 25, and the two resulting daughter cells are tracked as separate trajectories from frame 26 onwards, with both remaining stably tracked through frame 30. The mother track is correctly terminated at the point of division and two new daughter tracks are initialized, preserving the lineage relationship in the output data structure. No false splits were observed in the vicinity of this event, and both daughter trajectories were maintained without interruption for the remainder of the sequence.

\begin{figure}
\centerline{\includegraphics[width=\textwidth]{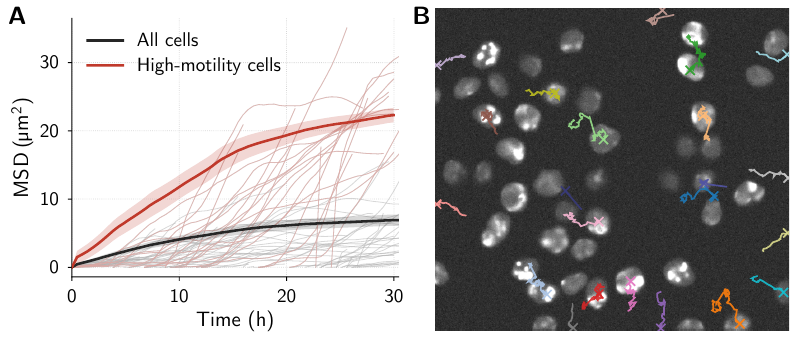}}
\caption{(A) Mean squared displacement (MSD) of cell trajectories from the Fluo-N2DH-SIM+ dataset over time. All tracked cell paths are shown, with those exceeding an MSD threshold of $1000~\mu\mathrm{m}^2$ highlighted in red. (B) Subset of trajectories corresponding to an MSD greater than $1000~\mu\mathrm{m}^2$ at the last time point.}
\label{fig3}
\end{figure}

Beyond trajectory generation, the framework enables quantitative analysis of cell migration behaviors. Figure~\ref{fig3}A shows mean squared displacement (MSD) analysis of tracked cells in the Fluo-N2DH-SIM+ dataset, highlighting heterogeneous migration phenotypes. MSD curves for all tracked cells reflect large heterogeneity in displacement magnitudes. Trajectories exceeding an MSD threshold of $1000~\mu\mathrm{m}^2$ are highlighted in red, identifying a subset of highly motile cells. Figure~\ref{fig3}B displays only these high-MSD trajectories, clearly delineating the actively migrating subpopulation from cells showing confined or subdiffusive motions.

\begin{figure}
\centerline{\includegraphics[width=\textwidth]{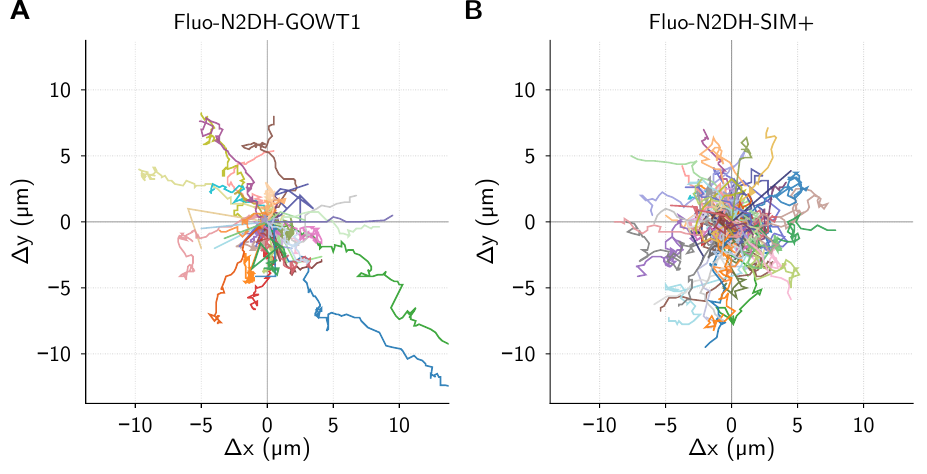}}
\caption{(A) Paths starting at (0,0) for datasets Fluo-N2DH-GOWT1 and (B) Fluo-N2DH-SIM+.}
\label{fig4}
\end{figure}

Figures \ref{fig4}A and \ref{fig4}B show normalized cell paths for Fluo-N2DH-GOWT1 and Fluo-N2DH-SIM+ datasets, where all trajectories originate at the coordinate origin to illustrate movement patterns independent of initial positions. The contrasting migration patterns are immediately apparent. Cells in Fluo-N2DH-SIM+ display random, non-directional motility with trajectories distributed isotropically around the origin, characteristic of Brownian-like displacements in spatially homogeneous environments. In contrast, Fluo-N2DH-GOWT1 cells exhibit more directional, persistent migration, with trajectories extending preferentially in certain directions, suggesting either chemotactic gradients or contact guidance mechanisms.

\section{Discussion}


We introduced ARGUS, a modular framework for cell tracking in time-lapse microscopy. The framework combines adaptive cell detection, Farneback optical-flow prediction, frame-to-frame linear assignment, and global tracklet refinement across short gaps. This decomposition allows each component to be independently adjusted or replaced. Users can modify detection parameters, adjust optical flow parameters for different movement patterns, or change association distance thresholds. Since the framework is not a black box, users can implement additional processing steps, incorporate domain-specific knowledge, or diagnose failure modes by examining which step causes problems. The simulated Fluo-N2DH-SIM+ dataset reached the highest accuracy (DET: 0.971, TRA: 0.964), followed by Fluo-C2DL-Huh7 (0.957/0.946) and Fluo-N2DH-GOWT1 (0.943/0.937). The phase-contrast PhC-C2DH-U373 dataset additionally required monogenic signal analysis with phase-symmetry filtering to suppress halo artifacts and remained the most challenging case (DET: 0.905, TRA: 0.897).

The significance of ARGUS lies in addressing a methodological gap that persists despite the high accuracy of deep learning methods on benchmark datasets. Although the top CTC performers reach DET and TRA scores near 0.999, their deployment requirements create barriers to adoption in many research settings: they typically depend on substantial annotated training data (often hundreds to thousands of cells) and GPU infrastructure, both of which can be prohibitive for resource-limited laboratories. ARGUS attains DET and TRA scores between 0.90 and 0.97 (above the CTC average on three of the four benchmark datasets) without requiring training data or GPU resources, and processes each frame in a few seconds. When ground-truth segmentation masks are supplied, tracking accuracy (TRA) rises to 0.999, indicating that the remaining error originates in cell detection rather than tracking and that the two stages can be evaluated, and improved, independently. This performance compares favorably with the 10–80\% accuracy typically achieved by manual tracking under time constraints, and it enables three deployment scenarios in which deep learning methods are impractical: (1) real-time monitoring of ongoing experiments, where rapid feedback is essential for adjusting experimental conditions; (2) exploratory studies of new cell lines or imaging conditions, where collecting training data is infeasible; and (3) generation of initial trajectory annotations that can be manually refined into training sets for machine learning methods. The modular architecture supports this incremental pathway, positioning ARGUS as both an immediate research tool and a transitional methodology: it can be applied directly with default parameters and then tuned iteratively from observed results, and for recurring protocols, optimized parameter sets can be saved and reused.

The computational efficiency of ARGUS scales favorably with dataset size and complexity. The initial tracker processes frame pairs sequentially, while the refinement adds a single assignment over tracklet endpoints rather than a detection-level graph over the full sequence. For typical datasets with 50--200 cells per frame and 100--300 frames, processing completes in under a minute on a standard workstation. Memory usage remains modest, and parallel optical-flow estimation supports batch processing across CPU cores. 

Beyond trajectory generation, ARGUS provides quantitative migration analysis. The MSD analysis (Figures \ref{fig3}A-B) identified two distinct populations in Fluo-N2DH-SIM+: most cells showed confined motion ($\mathrm{MSD}<1000~\mu\mathrm{m}^2$) while a subset showed ballistic behavior ($\mathrm{MSD}>1000~\mu\mathrm{m}^2$). This automated classification can distinguish quiescent versus actively migrating cells, or identify drug responders. The normalized path visualizations (Figures \ref{fig4}A-B) reveal dataset-specific patterns: Fluo-N2DH-SIM+ cells display random motion characteristic of homogeneous environments, while Fluo-N2DH-GOWT1 cells exhibit directed migration suggesting chemotaxis or contact guidance. The framework easily computes additional statistics including velocity distributions, directional persistence, and confinement ratios, that are the parameters of interest for biological investigations.

A typical workflow with ARGUS starts by applying the default preprocessing and detection parameters to a representative subset of frames in order to assess baseline performance. Users then visually inspect the detection results to verify accurate cell identification. Exhaustive parameter scans demonstrated that ARGUS requires little fine-grained tuning, as default parameter values transfer robustly across datasets with comparable cell sizes and imaging conditions. If detection accuracy remains poor, users can try different preprocessing strategies (monogenic filtering for phase contrast, wavelet denoising for high noise, or minimal preprocessing for clean fluorescence). Once detection is satisfactory, users run the full tracking pipeline. This staged design allows the pipeline to be readily adapted to previously unseen contrast mechanisms or cell types. Deployment of ARGUS as an executable application on a quantitative image-processing platform, such as the BIOQIC platform (\url{https://bioqic.charite.de/}), is currently pending. This represents an important next step toward broader usability, standardized access, and reproducible application by non-specialist users.

ARGUS prioritizes computational efficiency and interpretability while maintaining competitive accuracy. The global tracklet-refinement stage reduces fragmentation caused by short occlusions or missed detections, but it does not constitute fully global detection-level lineage optimization. Extended occlusions remain challenging, and the one-to-one endpoint assignment does not jointly optimize branching or fusion events. Divisions are therefore handled by the separate split-detection rule. Accuracy also depends on optical-flow quality, which may degrade under low contrast or rapid deformation, while extensive cell overlap may produce merged detections. A dedicated ablation isolating the optical-flow component from the linear-assignment step was not performed; however, the near-perfect tracking obtained with ground-truth masks (Table~\ref{tab:res}) suggests that, for the motion regimes represented in these datasets, flow-guided association is not the limiting factor. Despite these limitations, comparison with Trackastra showed equivalent linking performance when the same masks were supplied. The practical advantage of ARGUS is that it also produces these masks without training data.

Several extensions could address current limitations. Physical constraints from equations of motion, specifically conservation of mass and momentum, could be incorporated into the detection and tracking framework. The association step can be improved by these physics-based constraints through restricting trajectories to biologically and physically plausible paths, reducing false associations even during occlusions. To handle detection more robustly, adaptive thresholding could be replaced by learned segmentation networks (e.g., Cellpose, StarDist) while retaining the efficient optical-flow and linear-assignment stages. The modular structure creates a pathway toward hybrid pipelines combining algorithmic efficiency with learned components. Finally, the mathematical framework has potential to extend to 3D+t imaging. Three-dimensional thresholding could enable volumetric segmentation, while three-dimensional optical flow algorithms would enable tracking in organoids, embryos, and thick tissue sections. ARGUS represents both a practical tool for immediate application and a flexible platform for methodological innovation.

In summary, we introduced ARGUS, a training-free cell tracking framework designed for practical monitoring applications. ARGUS enables rapid, interpretable cell tracking and immediate deployment without training data or expensive GPU infrastructure. By combining cell detection, dense motion estimation, local association, and global tracklet refinement, the framework achieves detection accuracies of 0.905--0.971 and tracking accuracies of 0.897--0.964 in under one minute per dataset (5--6 seconds for 3 frames). This speed-accuracy trade-off makes ARGUS particularly suitable for high-throughput screening, real-time experiment monitoring, and exploratory studies where imaging conditions vary. Users can adaptively tune each processing step for their specific cell types, imaging modalities, and biological questions without retraining models. The modular architecture also allows integration of advanced components for users who later acquire training data, creating a pathway from quick monitoring tool to precision tracker. Beyond generating trajectories, the framework provides biological insights through migration analysis, mean squared displacement calculations, and movement phenotype classification. The algorithm is open-source at \url{https://github.com/Gitinc/argus}.

\appendix
\section{Data preparation modules}
\label{app:data_prep}

Data preparation is performed before threshold-based cell detection. For contrast enhancement, the working image channel was clipped at a dataset-dependent upper intensity percentile $p_c$ (Table~\ref{tab:params_runtime}). Pixel values above this percentile were set to the percentile value, and the clipped frame was then min--max normalized to the 8-bit range $[0,255]$. This percentile clipping compresses the influence of very bright pixels and increases the visibility of faint, low-intensity cells against the background.

After contrast enhancement, wavelet denoising was applied to the same image channel. In the implementation used here, this corresponds to single-channel wavelet shrinkage with the Daubechies-1 wavelet, soft thresholding, BayesShrink threshold selection, automatic noise-standard-deviation estimation, and an automatically selected decomposition level. The denoised output was rescaled back to the 8-bit range for the subsequent threshold-based detection step. These two operations were applied to every dataset because the ablation study showed that contrast enhancement is critical for detection and that wavelet denoising is necessary for noisy data without degrading lower-noise datasets.

Additional modules are selected only when the imaging modality or acquisition conditions introduce a specific artifact. Rigid registration can be used as a drift-correction step for time-lapse data affected by stage motion, using an intensity-based alignment method (e.g., elastix \cite{elastix}) to align each frame to a common reference coordinate system. Sliding-window mean subtraction can normalize temporal intensity variations, and unsharp masking \cite{Polesel_Ramponi_Mathews_2000} can sharpen cell boundaries.

For phase-contrast data, ARGUS uses monogenic phase-symmetry filtering to enhance cell bodies while suppressing halo artifacts. The monogenic signal extends the analytic signal to two dimensions, yielding a multi-component representation from which local amplitude, phase, and orientation can be derived \cite{Felsberg_Sommer_2002}. In our implementation, a single-scale log-Gabor bandpass filter (center wavelength 60 px, bandwidth parameter $\sigma_{\text{onf}} = 0.05$) \cite{bridge2017introduction} is applied in the frequency domain to obtain a bandpass version of the image, and the Riesz transform, the two-dimensional analog of the Hilbert transform \cite{Felsberg_Sommer_2002}, is then used to generate the quadrature responses. This yields three feature maps: an even-symmetric component $f_e(x)$ that responds to isotropic structures, and two odd-symmetric components $f_{o1}(x)$ and $f_{o2}(x)$ that are sensitive to directional contrasts. The phase-symmetry response is represented by the local energy
\begin{equation}
E(x) = f_e(x)^2 + f_{o1}(x)^2 + f_{o2}(x)^2,
\end{equation}
which is large at structured regions such as cell bodies and small in flat background. The energy map is smoothed with a Gaussian filter ($\sigma = 8~px$) to suppress residual noise and consolidate cell interiors. The resulting map emphasizes potential cell locations under low-contrast phase-contrast conditions.

\section*{Declaration of competing interest}
The authors declare that they have no known competing financial interests or personal relationships that could have appeared to influence the work reported in this paper.

\section*{Ethics approval and consent to participate}
Not applicable.

\section*{Data availability}
Data used in this study are available from the cited public Cell Tracking Challenge datasets. The code is available at \url{https://github.com/Gitinc/argus}.

\section*{Acknowledgements}
Funding by German Research Foundation (DFG) projects: CRC1340 Matrix in Vision, CRC1540 EBM, GRK2260 BIOQIC, FOR5628 and German Federation of Industrial Research Associations (AIF) project KK5611902 BM4 is greatly appreciated. This research was supported [in part] by the Intramural Research Program of the National Institutes of Health (NIH). The contributions of the NIH author(s) were made as part of their official duties as NIH federal employees, are in compliance with agency policy requirements, and are considered Works of the United States Government. However, the findings and conclusions presented in this paper are those of the author(s) and do not necessarily reflect the views of the NIH or the U.S. Department of Health and Human Services. These sponsors had no role in the study design, collection, analysis and interpretation of data, writing of this manuscript or the decision to submit the article for publication.

\section*{CRediT authorship contribution statement}
Noah Jaitner: Conceptualization, Methodology, Software, Validation, Formal analysis, Visualization, Writing -- original draft.\\
Kandice Tanner: Resources, Investigation, Writing -- review \& editing.\\
Ingolf Sack: Supervision, Funding acquisition, Writing -- review \& editing.\\
Hossein S.\ Aghamiry: Supervision, Conceptualization, Methodology, Writing -- original draft, review \& editing.

\bibliographystyle{unsrtnat}
\bibliography{biblio}

\end{document}